\documentclass[conference]{IEEEtran}
\IEEEoverridecommandlockouts

\usepackage{cite}
\usepackage{amsmath,amssymb,amsfonts}
\usepackage{algorithmic}
\usepackage[utf8]{inputenc}
\usepackage{graphicx}
\usepackage{textcomp}
\usepackage{xcolor}
\usepackage[inline]{enumitem}
\usepackage{booktabs}
\usepackage{multirow}
\usepackage{xcolor}
\usepackage{hyperref}
\usepackage{xurl}
\def\BibTeX{{\rm B\kern-.05em{\sc i\kern-.025em b}\kern-.08em
    T\kern-.1667em\lower.7ex\hbox{E}\kern-.125emX}}
\begin{document}

\title{Ontology-Guided, Hybrid Prompt Learning for Generalization in Knowledge Graph Question Answering}

\author{
\IEEEauthorblockN{Longquan Jiang\IEEEauthorrefmark{1}, Junbo Huang\IEEEauthorrefmark{1}, Cedric Möller\IEEEauthorrefmark{1}, and Ricardo Usbeck\IEEEauthorrefmark{2}}\\
\IEEEauthorblockA{\IEEEauthorrefmark{1}Department of Computer Science, University of Hamburg, Hamburg, Germany\\
Email: \{longquan.jiang, junbo.huang, cedric.moeller\}@uni-hamburg.de}
\IEEEauthorblockA{\IEEEauthorrefmark{2}Institute for Information Systems, Leuphana University Lüneburg, Lüneburg, Germany \\
Email: ricardo.usbeck@leuphana.de}
}

\maketitle

\begin{abstract}
Most existing Knowledge Graph Question Answering (KGQA) approaches are designed for a specific KG, such as Wikidata, DBpedia or Freebase. Due to the heterogeneity of the underlying graph schema, topology and assertions, most KGQA systems cannot be transferred to unseen Knowledge Graphs (KGs) without resource-intensive training data. 
We present OntoSCPrompt, a novel Large Language Model (LLM)-based KGQA approach with a two-stage architecture that separates semantic parsing from KG-dependent interactions. OntoSCPrompt first generates a SPARQL query structure (including SPARQL keywords such as SELECT, ASK, WHERE and placeholders for missing tokens) and then fills them with KG-specific information. To enhance the understanding of the underlying KG, we present an ontology-guided, hybrid prompt learning strategy that integrates KG ontology into the learning process of hybrid prompts (e.g., discrete and continuous vectors). We also present several task-specific decoding strategies to ensure the correctness and executability of generated SPARQL queries in both stages. 
Experimental results demonstrate that OntoSCPrompt performs as well as SOTA approaches without retraining on a number of KGQA datasets such as CWQ, WebQSP and LC-QuAD 1.0 in a resource-efficient manner and can generalize well to unseen domain-specific KGs like DBLP-QuAD and CoyPu KG \footnote{Code: \href{https://github.com/LongquanJiang/OntoSCPrompt}{https://github.com/LongquanJiang/OntoSCPrompt}}.
\end{abstract}


\section{Introduction}
\begin{figure*}
    \centering
    \includegraphics[width=\linewidth]{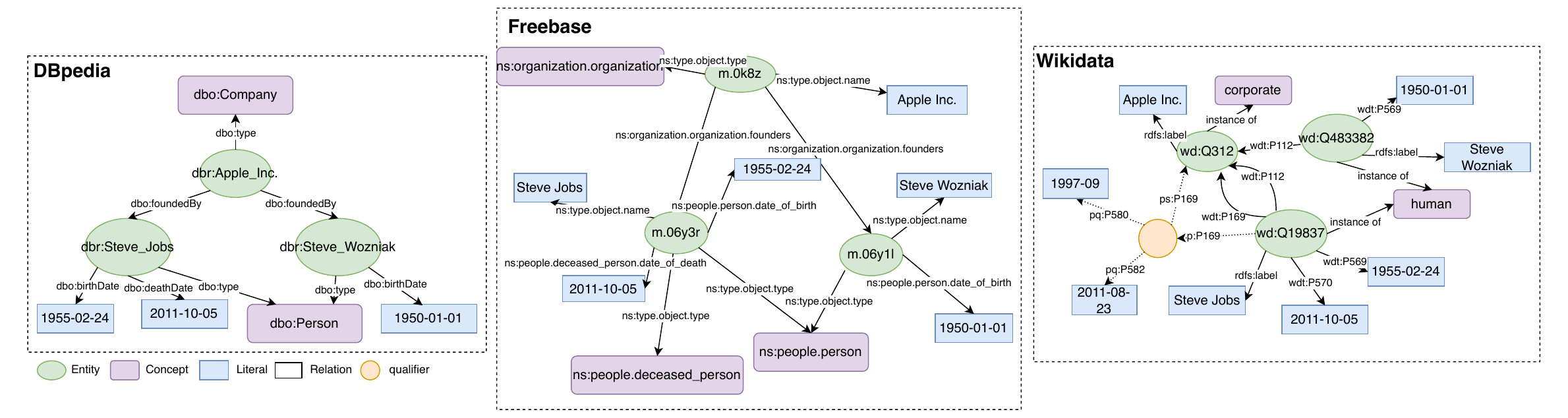}
    \caption{Three ontology examples representing the same world facts about \textit{Apple Inc.}, \textit{Steve Jobs} and \textit{Steve Wozniak} in Freebase, DBpedia and Wikidata. Similar knowledge can be modelled differently regarding assertions (i.e., persistent entity identifiers), schema and topology, requiring different translations from the same natural language question to a SPARQL query.\\}
    \label{fig:example}
\end{figure*}

KGQA systems enable non-expert users to pose natural language queries and retrieve precise and relevant answers from the underlying KG based on the facts available in the KG. There's a significant need for a KGQA system that can generalize across diverse KGs. This is a challenging task due to the heterogeneity of the underlying KG. As shown in Figure~\ref{fig:example}, Freebase, DBpedia and Wikidata, the most popular three general KGs, have their unique way of representing the same world facts regarding \textit{Apple Inc.}, \textit{Steve Jobs} and \textit{Steve Wozniak}. Heterogeneity can be found as \begin{enumerate*}[label=\arabic*)]
    \item \textbf{schema heterogeneity\footnote{We use schema and ontology or T-Box interchangeably.}}: differences in the concepts\footnote{Here, concepts and RDF classes are equivalent.} and the relations between them across different KGs. For instance, the three KGs in Figure~\ref{fig:example} have their own naming convention for the same concept of a \textit{Person}, namely \textit{dbo:Person} in DBpedia, \textit{ns:people.person} in Freebase, and \textit{human} in Wikidata;
    \item \textbf{topology heterogeneity}: differences in how a fact is represented and accessed within a KG. For instance, Wikidata utilizes a special type of connection known as a ``qualifier'' (depicted as nodes in orange in Figure~\ref{fig:example}) to furnish the triple \textlangle \textit{wd:Q19837, wdt:P169, wd:Q312}\textrangle with additional details like the start date;
    \item \textbf{assertions heterogeneity}: differences in the assertion about entities and their relations across different KGs. For instance, the assertion ``Joe Biden is the President of the United States'' is represented as \textlangle Joe Biden, office, President of the United States\textrangle in DBpedia, whereas in Wikidata, it is \textlangle Joe Biden, position held, President of the United States\textrangle.
\end{enumerate*}

The majority of current KGQA systems lack generalization because they are typically tailored for a particular KG~\cite{kapanipathi-etal-2021-leveraging,10.1145/2588555.2610525,10.1145/3357384.3358026}, or focus only on within-a-KG generalization~\cite{10.1145/3442381.3449992,gu-etal-2023-dont,shu-yu-2024-distribution}. Although some approaches~\cite{sun-etal-2018-open,saxena-etal-2020-improving,qiao-etal-2022-exploiting,mavromatis-karypis-2022-rearev} showed their ability to generalize across KGs in certain respects, e.g., regarding the assertion heterogeneity which lies between dataset identifiers such as WebQSP (Freebase)~\cite{10.1145/1376616.1376746} and MetaQA (Wikimovies)~\cite{10.5555/3504035.3504780}, they fail to generalize to other aspects such as different schemas or topologies.

Large language models (LLMs) have demonstrated remarkable reasoning capabilities. However, several studies reveal that LLMs perform inadequately in knowledge-intensive task KGQA~\cite{2023arXiv230310368H,2023arXiv230307992T,2023arXiv230911206W}. LLMs suffer from issues such as hallucination~\cite{10.1145/3571730} and factual inaccuracy when answering questions, mainly because they lack domain-specific knowledge, stemming from insufficient training data or missing interactions with an unseen or heterogeneous KG \cite{DBLP:conf/esws/KlagerP23}.

We present \textbf{OntoSCPrompt}, a two-stage ontology-guided hybrid prompt learning KGQA method. To be agnostic of the underlying KG, we use a two-stage process~\cite{ravishankar-etal-2022-two,10.1145/3589292} that separates semantic parsing from intensive KG-dependent interactions. First, we forecast a \textit{SPARQL Query Structure} independent of any specific KG. Second, we fill the placeholders with missing KG identifiers, such as entities and relations. To enhance the understanding of the semantics of the underlying KG, we integrate ontology knowledge into the learning process of hybrid prompts, i.e., discrete and continuous vectors, in both stages. We develop several task-specific decoding strategies to ensure the validity of the generated SPARQL queries (grammar and structure). Our evaluation uses KGQA datasets derived from heterogeneous KGs, such as Freebase~\cite{10.1145/1376616.1376746}, DBpedia~\cite{10.1007/978-3-540-76298-0_52} and DBLP~\cite{DBLP:conf/birws/BanerjeeAUB23}.

Our main contributions are as follows:
\begin{enumerate*}[label=\arabic*)]
    \item A novel two-stage KGQA system that can generalize across multiple KGs with a KG ontology-guided hybrid prompt learning strategy.
    \item To the best of our knowledge, our approach is the first to apply prompt tuning to the KGQA generalization and evaluation task. Experimental results demonstrate its effectiveness in understanding the semantics of the underlying KG and facilitating knowledge transfer across multiple KGs.
    \item We design several decoding strategies tailored to our two-stage approach, e.g., grammar-constrained and structure-guided techniques, to ensure the validity of generated SPARQL queries, thereby closely connecting the individual modules and enhancing generalization.
\end{enumerate*}

\section{Methodology}

In this section, we explain the two-stage approach \textbf{OntoSCPrompt} and how it facilitates generalization across diverse KGs.

\subsection{Two-Stage Framework}
\label{sec:divide}

As discussed earlier, a KGQA system that can generalize to unseen KGs is expected to comprehend the similarities and differences between KGs. That is, for the same natural language question and a different KG on which it is trained, the system should be able to generate a matching SPARQL query without extensive retraining. To this end, we utilize a two-stage framework: \begin{enumerate*}[label=\arabic*)]
    \item \textbf{Query Structure Prediction}, wherein questions are translated into generic SPARQL query structures independent of any particular KG, called the structure stage (Stage-S).
    \item \textbf{KG Content Population}, where the SPARQL structures are populated with schema elements such as concepts, relations and entities specific to the given KG, called the content stage (Stage-C). 
\end{enumerate*} Finally, the SPARQL query for a given question is generated by combining the predictions of these two sub-tasks.

\subsection{Structured SPARQL Query}

The inherent heterogeneity of the schema of different KGs, including entity identifiers, concepts, and relations, challenge KGQA systems to adapt their understanding and reasoning processes across different KGs. To tackle this challenge, we design a \emph{generic query structure representation}, which consists of the following elements of standard SPARQL queries: 1) \textbf{Reserved Keywords}, such as SELECT, ASK, FILTER, COUNT, etc.; 2) \textbf{Literals}, such as numbers, strings, dates, and fixed textual values; 3) \textbf{KG-specific Identifiers}, such as entities, relations, concepts, and variables. Many linguistically similar questions share similar SPARQL skeletons, even across different KGs, if not identical, due to ontological modelling based on humans' natural language usage.

We extend previous work~\cite{DBLP:conf/semweb/00010U23} by adding a new placeholder for concepts and supporting more complex SPARQL clauses like having, group by, order by, etc. We also prove that this approach with our extensions can generalize well to other KGs without retraining. We use 6 special tokens to serve as placeholders for to-be-filled components within SPARQL queries. These tokens are: \begin{enumerate*}[label=\arabic*)]
    \item \textit{[ent]} for entities mentioned in a given question, 
    \item \textit{[cct]} for concepts, 
    \item \textit{[rel]} for relations specified in a KG ontology, 
    \item \textit{[var]} for variables,  
    \item \textit{[val]} for literals appearing in basic graph patterns\footnote{\url{https://www.w3.org/2001/sw/DataAccess/rq23/\#BasicGraphPatternMatching}}, value clauses or solution modifiers of SPARQL queries, and
    \item \textit{[con]} to signify a constraint or condition in such SPARQL keywords as FILTER, ORDER BY, GROUP BY or HAVING. Such a placeholder enhances cross-KG alignment and broadens the coverage of complex questions with constraint(s).
\end{enumerate*}

In terms of reasoning difficulty, our structure representation can cover structures of questions that require single-hop or multi-hop reasoning (with or without aggregates, conditions or both). Table~\ref{tab:query-complexity} shows examples from simple questions to complex questions with or without constraints and/or aggregates. We will use this structure in \textbf{Stage-S} to guide the LLM while constructing the SPARQL query. 

\begin{table}[tbh!]
\caption{Examples of SPARQL query structures supported by our proposed method.}
\footnotesize
\centering
\begin{tabular}{|p{0.3\textwidth}|p{0.1\textwidth}|l|c|}
\hline
 \textbf{SPARQL Query Structure}  & \textbf{Type} \\
       \hline
      select [var] where \{ [ent] [rel] [var] \}                &        Single-hop                    \\
      \hline
      ask where \{ [ent] [rel] [ent] \}           &        Single-hop                   \\
      \hline
     select ( count ( [var] ) as [var] ) where \{ [ent] [rel] [var] \}                 &      Single-hop with aggregate                  \\
     \hline
    select [var] where \{ [ent] [rel] [var] . [ent] [rel] [cct] . \}              &         Multi-hop               \\
\hline
select ( min ( [var] ) as [var] ) where \{ [var] [rel] [var] . [ent] [rel] [var] . \}              &      Multi-hop with aggregates     \\
\hline
select [var] where \{ [var] [rel] [var] . [ent] [rel] [var] . filter [con] \} & Multi-hop with constraints \\
\hline
select ( count ( [var] ) as [var] ) where \{ [var] [rel] [var] . [ent] [rel] [var] . filter [con] \} & Multi-hop with constraints and aggregates \\
\hline

\end{tabular}

\label{tab:query-complexity}
\end{table}


\subsection{Ontology-Guided Hybrid Prompt Learning}
\label{sec:prompt}

Our prompt construction offers LLMs with task-specific information for accurate predictions. There are two prompt construction methods: (1) \textbf{Textual Prompts}, a textual template like "Answer the question: [input], [output]" to guide LLMs to generate the desired output; (2) \textbf{Learnable Vectors}, a series of continuous vectors prepended to the input which can be optimized during training. Here, we have two main considerations in prompt construction: (1) to mitigate the effect caused by the heterogeneity of the underlying KGs and (2) to ensure the adaptability to new KGs. Thus, we propose a novel ontology-guided, hybrid prompt learning method.

\paragraph{Task-specific, Ontology-Guided Textual Prompts.} Ontology knowledge is explicitly prepended to enhance the understanding of KG-specific semantics in two stages. The structure prompt $P_s$ in \textbf{Stage-S} and content prompt $P_c$ in \textbf{Stage-C} are designed as follows. 

\begin{itemize}
\small
    \item $P_s = $ "[prefix] [question] | [ontology]" 
    \item $P_c = $ "[prefix] [structure] | [question] | [ontology]"
\end{itemize}

We use "translate the question into sparql according to the ontology:" as [prefix]. We use a verbalization method~\cite{text2kgbench} to convert the ontology to text format. For example, the DBpedia ontology in Figure~\ref{fig:example} is verbalized as "ontology: concepts: Company, Person; relations: foundedBy, birthDate, deathDate, type; entities: Steve\_Jobs, Steve\_Wozniak, Apple\_Inc."

\paragraph{Aspect-aware Continuous Prompts.} We introduce four learnable vectors~\cite{10.1145/3589292} with random initialization for understanding different aspects of the input, i.e., $\mathbf{v}^{Q}$ for learning question $Q$ specific features, $\mathbf{v}^{\mathcal{G}}$ for learning ontology $\mathcal{G}$ specific features, $\mathbf{v}^{B}$ and $\mathbf{v}^{E}$ for learning task-specific features at the beginning and end of the input respectively (see Equation~\ref{eqn1} and \ref{eqn2}). 
It holds that $\mathbf{v}^{Q},\mathbf{v}^{\mathcal{G}},\mathbf{v}^{E}, \mathbf{v}^{B} \in \mathbb{R}^d$ where $d$ is the dimensionality of the LLM input representations.

Therefore, hybrid prompts (i.e., prompts containing textual and continuous parts) for \textbf{stage-S} and \textbf{stage-C}, i.e., $\mathbf{I}_S$ and $\mathbf{I}_C$ are constructed as follows:

\begin{align}
  \mathbf{I}_S &= \mathbf{v}^B \oplus \mathbf{e}^S \oplus \mathbf{v}^Q \oplus \mathbf{e}^Q \oplus \mathbf{v}^\mathcal{G} \oplus \mathbf{e}^\mathcal{G} \oplus \mathbf{v}^E \label{eqn1} \\
  \mathbf{I}_C &= \mathbf{v}^B \oplus \mathbf{e}^C \oplus \mathbf{v}^Q \oplus \mathbf{e}^Q \oplus \mathbf{v}^\mathcal{G} \oplus \mathbf{e}^\mathcal{G} \oplus \mathbf{v}^E \label{eqn2}
\end{align}

where, $\mathbf{e}^{S}$, $\mathbf{e}^{C}$, $\mathbf{e}^{Q}$ and $\mathbf{e}^{\mathcal{G}}$ represent the embedding of structure prefix in $P_S$, content prefix in $P_C$, question $Q$ and ontology $\mathcal{G}$ respectively. Here, $\mathbf{e}^{S},\mathbf{e}^{C},\mathbf{e}^{Q}, \mathbf{e}^{\mathcal{G}} \in \mathbb{R}^d$ and $\oplus$ is the operator for vector concatenation. 

\paragraph{Hybrid Prompt Learning.} At both stages, the trainable parameters $\Theta$ correspond to the base model parameters $\Theta_{m}$ and the learnable vectors $\Theta_{l}$. The desired output is generated through auto-regressive decoding\cite{10.1145/3589292,zheng-lapata-2021-compositional-generalization}. The parameter $\Theta$ is learnt by minimizing the negative log-likelihood loss.
\begin{equation}
\small
\begin{split}
   L(\Theta) &= - \frac{1}{n} \sum_{i=1}^{n} \log P(O_{gold}^{i} | \mathbf{I}^{i}; \Theta) \\
   &= -\frac{1}{n} \sum_{i=1}^n \sum_{j=1}^{m_i} \log P(O_{gold}^{i,j} | \mathbf{I}^i ; O_{gold}^{i,1},...,O_{gold}^{i,j-1}; \Theta) \\
   \end{split}
\end{equation}

where $\mathbf{I}$ and $O$ are the hybrid prompts as the input and the ground truth output at each stage. 

For example, given a question, ``Who founded Apple?'', the ground truth output $O_S$ is its SPARQL structure ``select [var] where \{ [ent] [rel] [var] \}'' at the structure stage while the ground truth output $O_C$ is "[var] var0 [ent] dbr:Apple\_Inc. [rel] dbo:foundedBy [var] var0"  at the content stage.


\subsection{Constrained Decoding Strategies}
\label{sec:decoding}

Constrained decoding approaches facilitate the generation of text sequences in a controllable and expected fashion. This technique is widely used in neural machine translation~\cite{leblond-etal-2021-machine}, text summarization~\cite{fan-etal-2018-controllable}, and neural semantic parsing~\cite{baranowski-hochgeschwender-2021-grammar2}. 
To ensure the validity of the generated SPARQL queries, thus, we devise different task-specific decoding strategies: \begin{enumerate*}[label=\arabic*)]
    \item Grammar-constrained Decoding at \textbf{stage-S}, and 
    \item Structure and/or Subgraph Guided Decoding at \textbf{stage-C}.
\end{enumerate*}

\textbf{Grammar-constrained Decoding}. Integrating grammar constraints at the decoding stage guarantees grammatical correctness, particularly in low-resource situations~\cite{geng-etal-2023-grammar}. For example, the model must have the ability to discard outputs like "select [var] where \{ [var] [ent] [ent] \}", as they do not conform to our simplified SPARQL grammar rules where we expect a [rel] instead of [ent]. Referring to the standard SPARQL grammar definition\footnote{\url{https://www.w3.org/TR/2013/REC-sparql11-query-20130321/\#sparqlGrammar}}.

\textbf{Structure-guided Decoding}. To ensure the validity of the resulting SPARQL query, the content predicted in \textbf{Stage-C} must be consistent with the placeholders predicted in \textbf{Stage-S}. In the beam search process, we adjust the score of the candidate placeholder tokens - which do not align with their respective counterparts in the structure - to $-\infty$. For example, if the predicted structure is "select [var] where \{ [ent] [rel] [var] \}", and the predicted content is "[var] var0 [var] var0 [rel] dbo:founders [ent] dbr:Microsoft", the model fails to merge them into the final query due to structure inconsistency.


\textbf{Subgraph Constrained Decoding}. Subgraphs of question entities provide contextual information. 
Understanding the surrounding context helps to disambiguate the meaning of entities or relations~\cite{DBLP:conf/acl/ZhangZY000C22,jiang-etal-2023-reasoninglm}. Thus, we introduce subgraph constraints in stage-C to assign higher priority to relations in a subgraph relevant to the given question. 
Consider the question "Who plays Ray Barone?" in the WebQSP dataset. It is challenging for the model to differentiate between "film.performance.actor" and "tv.regular\_tv\_appearance.actor". Subgraph constraints prefer to choose "tv.regular\_tv\_appearance.actor" as "film.performance.actor" does not exist in the extracted subgraph of the entity "m.05h7f2 (Ray Barone)". 

\section{Experimental Setup} 

\subsection{Datasets}
We use datasets across a wide range of KGs, such as Freebase, DBpedia and DBLP, to show the generalization abilities of OntoSCPrompt: \begin{enumerate*}[label=\arabic*)]
    \item \textbf{WebQSP} (Freebase): A popular dataset with 4,937 questions extracted from Google Search logs. Those questions involve up to 2-hop reasoning and constraints. 
    \item \textbf{LC-QuAD 1.0} (DBpedia): The dataset consists of question and SPARQL query pairs generated using predefined question templates and crowdsourcing. It contains diverse types of complex questions, such as simple, multi-hop, and aggregation.
    \item \textbf{CWQ} (Freebase): A KGQA benchmark modified from WebQSP dataset, having a higher percentage of complex questions with multi-hops and constraints. 
    \item \textbf{SimpleDBpediaQA} (DBpedia): A mapping of the SimpleQuestions dataset from Freebase to DBpedia. 
    \item \textbf{DBLP-QuAD} (DBLP): A newly released complex question benchmark over the scholarly KG (i.e., DBLP) with 10,000 pairs of question and SPARQL queries. 
    \item \textbf{CoyPuKGQA} (CoyPu KG): a newly created KGQA benchmark over an industrial, global economy KG, namely the CoyPu KG, with 939 questions\footnote{\url{https://github.com/semantic-systems/coypu-KGQA-Dataset}}.
\end{enumerate*} Table~\ref{tab:statistics-datasets} shows the statistics of the datasets and Table~\ref{tab:statistics-structures} shows the statistics of SPARQL query structures in each dataset and the questions which have unseen query structures in the test set.

\begin{table}[htb!]
\caption{Dataset statistics.}
\small
\centering
\begin{tabular}{|c|c|c|c|}
\hline
       & \textbf{Train} & \textbf{Valid} & \textbf{Test} \\
       \hline
WebQSP          &      3,098        & -              &        1,639                    \\
CWQ             &      27,639       &       3,519       &        3,531                   \\
DBLP-QuAD       &      7,000        &      1,000        &      2,000                  \\
LC-QuAD 1.0     &      4,000        &     -          &         1,000               \\
SimpleDBpediaQA &       30,186        &       4,305        &      8,595                \\
CoyPuKGQA &       873        &       -        &      66                \\
\hline
\end{tabular}
\label{tab:statistics-datasets}
\end{table}

\begin{table}[htb!]
\caption{The number of unique SPARQL query structures. The column "\#S" represents the amount of structures only present in the test set but absent in the train set. The column "\#Q" represents the number of questions in the test set whose structure is unseen in the train set.}
\resizebox{\linewidth}{!}{
\centering
\begin{tabular}{|c|c|c|c|c|c|}
\hline
       & \textbf{Train} & \textbf{Valid} & \textbf{Test} & \textbf{\#S} & \textbf{\#Q} \\
       \hline
WebQSP          & 73             & -              & 53            & 18                  &     21                \\

CWQ             & 267            & 78             & 105           & 13                  &     31                \\

DBLP-QuAD       & 56             & 64             & 65            & 9                   &      193               \\

LC-QuAD 1.0     & 22             & -              & 21            & 0                   &      0               \\

SimpleDBpediaQA & 4              & 4              & 4             & 0                   &     0       \\
CoyPuKGQA & 48              & -              & 44             & 3                   &     5       \\
\hline

\end{tabular}
}
\label{tab:statistics-structures}
\end{table}

\subsection{Baselines}

We compare OntoScPrompt to several existing KGQA systems, which were themselves evaluated over different KG sets: \begin{enumerate*}[label=\arabic*)]
    \item \textbf{STaG-QA}~\cite{ravishankar-etal-2022-two} is a KGQA system for evaluating generalizability on WebQSP, LC-QuAD 1.0, MetaQA and SimpleQuestionWikidata, separating the cross-KG reasoning process into two stages, i.e., softly-tied query sketch generation and KG alignment. 
    \item \textbf{GraphNet}~\cite{sun-etal-2018-open} is a method that integrates information from both knowledge bases and text corpora at an early stage of processing. 
    \item \textbf{PullNet}~\cite{sun-etal-2019-pullnet} use an iterative process to build a question-specific subgraph. In each iteration, a graph-CNN is used to pinpoint subgraph nodes that should be expanded.
    \item \textbf{EmbedKGQA}~\cite{saxena-etal-2020-improving} leverages KG embeddings to perform multi-hop KGQA on WebQSP and MetaQA datasets.
    \item \textbf{HGNet}~\cite{10.1109/TKDE.2022.3207477} is an end-to-end method for query graph generation, using hierarchical autoregressive decoding and a unified graph grammar AQG to delineate the structure of query graphs.
    \item \textbf{TERP}~\cite{qiao-etal-2022-exploiting} integrates explicit textual information and implicit KG structural features based on a novel entity link prediction framework.
\end{enumerate*}

\subsection{Evaluation Metrics}

For comparison, we use Precision, Recall and F1 as standard evaluation metrics for LC-QuAD 1.0, SimpleDBpediaQA and DBLP-QuAD, and Hits@1 for CWQ and WebQSP. Note that the Exact Match (EM) score metric is used while training, as the target in each stage, is either the structure or content of a SPARQL query instead of full SPARQL queries.

\subsection{Implementation Details}

\textit{KG Endpoints}. We use DBpedia 2016-10\footnote{\url{https://downloads.dbpedia.org/2016-10/}} for LC-QuAD 1.0 and SimpleDBpediaQA, the latest Freebase dump\footnote{\url{https://developers.google.com/freebase}} for WebQSP and CWQ. We host local SPARQL Virtuoso endpoints for DBpedia and Freebase. For querying the DBLP KG, we use the official live SPARQL query endpoint\footnote{\url{https://dblp-kg.ltdemos.informatik.uni-hamburg.de/sparql}} for DBLP-QuAD \cite{DBLP:conf/birws/BanerjeeAUB23}.

\textit{Data preprocessing}. We preprocess~\cite{DBLP:conf/semweb/00010U23} the SPARQL queries in each benchmarking dataset, which involves prefix removal, variable name standardization, lowercasing, redundant whitespace removal, prefixing IRIs and so on. Note that the preprocessing procedure does not change the semantics or executability of the SPARQL queries. For example, the resource "\textlangle http://dbpedia.org/resource/Microsoft\textrangle" is replaced with "dbr:Microsoft". We then split each ground truth SPARQL query into two parts, i.e., structure and content. The structure part is the query whose all schema elements (e.g., entities, relations, etc.) are replaced with the predefined placeholders. The content part is the concatenation of each placeholder and its corresponding schema element. Finally, we perform a consistency check to ensure that the preprocessed SPARQL queries can be restored by merging their corresponding structure and content.

\textit{Model configuration and parameters}. We use LongT5~\cite{guo-etal-2022-longt5} as our base model, and its publicly available Huggingface implementation\footnote{\url{https://huggingface.co/google/long-t5-local-base}}. Following Gu et al.~\cite{10.1145/3589292}, we use Adafactor to optimize the parameters in our proposed model. To enhance training stability across KGQA datasets, we first set a learning rate of 0.1 to train the learnable vectors and then set a learning rate of 5e-5 to train both the learnable vectors and the base model. For subgraph retrieval, we use the subgraph retriever~\cite{zhang-etal-2022-subgraph}, where the subgraph is induced by expanding top-K paths relevant to the given question from the topic entities, and set TOP\_K to 20 and min\_score to 1e-5.

\section{Evaluation}

In this section, we examine our experimental findings and assess how effective OntoSCPrompt is for KGQA generalization. First, we gauge OntoSCPrompt's performance on individual KGQA datasets. 
Second, we assess OntoSCPrompt's capacity to generalize across KGQA datasets within the same KG. 
Third, we evaluate its capability to generalize across different KGs.

\subsection{Evaluation on Generalization}

\begin{table*}[tbh!]
\caption{Comparison with prior state-of-the-art methods. We report F1 score on LC-QuAD 1.0, Hits@1 on WebQSP. "-" indicates no result reported on this dataset.}
\resizebox{\textwidth}{!}{
\centering
\begin{tabular}{|c|c|c|c|c|ccc|}
\hline
 \multirow{2}{*}{\textbf{Systems}}    &  \multirow{2}{*}{\textbf{Heterogeneity}} &   \multirow{2}{*}{\textbf{KG(s)}}   & \textbf{WebQSP} & \multicolumn{3}{c|}{\textbf{LC-QuAD 1.0}}   \\ 
  &    &      & Hits@1 & \multicolumn{1}{c}{P}    & \multicolumn{1}{c}{R}    & \multicolumn{1}{c|}{F1}   \\ 
\hline
EmbedKGQA & Assertion  & Freebase, MetaQA        &     66.6            & \multicolumn{1}{c}{-}    & \multicolumn{1}{c}{-}    & \multicolumn{1}{c|}{-}    \\ 
GraftNet   & Assertion &  Freebase, MetaQA  & 67.8            & \multicolumn{1}{c}{-}    & \multicolumn{1}{c}{-}    & \multicolumn{1}{c|}{-}    \\ 
PullNet  & Assertion &   Freebase, MetaQA     & 68.1            & \multicolumn{1}{c}{-}    & \multicolumn{1}{c}{-}    & \multicolumn{1}{c|}{-}    \\
TERP  & Assertion &   Freebase, MetaQA     & 76.8            & \multicolumn{1}{c}{-}    & \multicolumn{1}{c}{-}    & \multicolumn{1}{c|}{-}    \\
\hline
HGNet  &  Topology, Schema  &  Freebase, DBpedia         & 70.6           & \multicolumn{1}{c}{75.8}    & \multicolumn{1}{c}{\textbf{75.2}}    & \multicolumn{1}{c|}{75.1}    \\ 
STaG-QA &  Topology, Schema   &  Freebase, DBpedia, WD, MetaQA         & 65.3            & \multicolumn{1}{c}{76.5} & \multicolumn{1}{c}{52.8} & \multicolumn{1}{c|}{51.4} \\ 
OntoSCPrompt w/o constraints   &  Topology, Schema  & \multirow{2}{*}{Freebase, DBpedia, DBLP, CoyPuKG}         & 65.5            & \multicolumn{1}{c}{84.3} & \multicolumn{1}{c}{60.1} & \multicolumn{1}{c|}{70.2} \\ 
OntoSCPrompt with constraints   &  Topology, Schema  &        & \textbf{73.8 }           & \multicolumn{1}{c}{\textbf{92.9}} & \multicolumn{1}{c}{68.8} & \multicolumn{1}{c|}{\textbf{79.1}} \\ 
\hline
\end{tabular}
}

\label{tab:overall}
\end{table*}

Table~\ref{tab:overall} and Table~\ref{tab:evaluation_2} show the overall results of OntoSCPrompt on WebQSP, CWQ, LC-QuAD 1.0 and SimpleDBpediaQA in comparison to baselines for KGQA generalization. We train and evaluate each individual dataset based on its respective KG. 

Our proposed method, OntoSCPrompt, demonstrates competitive or state-of-the-art accuracy on both LC-QuAD 1.0 and CWQ compared to existing KGQA generalization baselines. From Table~\ref{tab:overall}, we observe that (1) OntoSCPrompt achieves an F1 score of 79.1\% on LC-QuAD 1.0, surpassing STaG-QA and HGNet by significant margins, namely \textbf{35.0\%} and \textbf{5.1\%} respectively. (2) OntoSCPrompt performs better than HGNet and STaG-QA on WebQSP by \textbf{4.5\%} and \textbf{13\%}. However, it underperforms TERP by \textbf{4.1\%}. The main reason is that TERP exploited relation paths’ hybrid semantics (explicit text information and implicit KG structural features). (3) Constrained decoding contributes significantly to OntoSCPrompt's overall performance improvement on WebQSP and LC-QuAD 1.0, respectively. (4) The methods addressing schema or topology heterogeneity demonstrate relatively lower performance compared to those targeting assertion heterogeneity, such as TERN, GrapftNet, PullNet and EmbedKGQA, since schema or topology heterogeneity is more complicated than assertion heterogeneity.

\begin{table}[]
\caption{The results on both CWQ and SimpleDBpedia for generalization within the same KG.}
\resizebox{\linewidth}{!}{
\centering
\begin{tabular}{|c|c|c|c|c|}
\hline
 \textbf{Systems} & \textbf{Trained} & \textbf{Tested} & \textbf{CWQ} & \textbf{SimpleDBpedia} \\ 
 \hline
EmbedKGQA & $\mathcal{D}^{A}$ & $\mathcal{D}^{A}$ & 44.7 & - \\  
PullNet & $\mathcal{D}^{A}$  &  $\mathcal{D}^{A}$  & 45.9 & - \\
TERP & $\mathcal{D}^{A}$  &  $\mathcal{D}^{A}$  & 49.2 & - \\
HGNet & $\mathcal{D}^{A}$  & $\mathcal{D}^{A}$ & 58.1 & - \\  
\hline
\multirow{2}{*}{OntoSCPrompt}  & $\mathcal{D}^{B}$, $\mathcal{G}^{B}$ & $\mathcal{D}^{A}$, $\mathcal{G}^{A}$ & 48.8 & 34.0 \\ 
 & $\mathcal{D}^{A}$, $\mathcal{G}^{A}$ &  $\mathcal{D}^{A}$, $\mathcal{G}^{A}$ & \textbf{70.4} & 84.6 \\ 
\hline
\end{tabular}
}
\label{tab:evaluation_2}
\end{table}



\subsubsection{Generalization Within the Same KG}

KGQA generalization within the same KG refers to the ability of a QA system to provide accurate answers to questions across different subsets or versions of the same KGQA dataset. Essentially, it involves the transfer of learned knowledge and reasoning capabilities within a single KG. To assess the ability of OntoSCPrompt, without any fine-tuning, we directly evaluate the model trained on the source KGQA dataset on the target KGQA dataset. We assume the source and target KGQA datasets are heterogeneous, as they handle different subsets of the same KG despite overlapping schema elements.

Table~\ref{tab:evaluation_2} shows the performance on CWQ using the WebQSP-trained model and the performance on SimpleDBpediaQA using the LC-QuAD 1.0-trained model. Following previous works, we report Hits@1 for CWQ and F1 score for SimpleDBpedia. $\mathcal{D}^{x}$ and $\mathcal{G}^{x}$ indicate the data split of the dataset $x$ on which the model is trained and test, with the ontology of the dataset $x$ integrated respectively. $A$ is the target dataset, i.e., CWQ or SimpleDBpedia, $B$ is the source dataset, i.e., WebQSP for CWQ or LC-QuAD 1.0 for SimpleDBpedia. OntoSCPrompt achieves Hits@1 of 48.8\% and F1 score of 34.0\% respectively, without any fine-tuning, only with their ontology provided, which performs below those trained and evaluated on the source KGQA dataset, such as TERP and HGNet. However, OntoSCPrompt achieves Hits@1 of 70.4\% and performs above TERP by 43\%, HGNet by 21.2\%, PullNet by 53.4\%, and EmbedKGQA by 57.5\% respectively, if fine-tuned on the target KGQA dataset with its ontology, i.e., trained on $\mathcal{D}^{A}$, $\mathcal{G}^{A}$. For SimpleDBpediaQA, we can also see the performance gain brought by KG ontology and fine-tuning. Thus, we demonstrate OntoSCPrompt's potential to generalize across different KGQA datasets within the same KG, even without any fine-tuning. This highlights the importance of the ontology on understanding the semantics of the underlying KG. 
Remember, the model has never seen any of these datasets (see trained on $\mathcal{D}^{B}$, $\mathcal{G}^{B}$ and tested on $\mathcal{D}^{A}$, $\mathcal{G}^{A}$). From earlier papers, we know that other models do not achieve any hits~\cite{hartmann-marx-soru-2018}.


\subsubsection{Generalization Across Different KGs}

\begin{table}[htb!]
\caption{F1 scores on DBLP-QuAD and CoyPuKGQA. $pre$ indicates the "pre-trained" variant of OntoSCPrompt using LC-QuAD 1.0 dataset.}
\small
\centering
\begin{tabular}{|c|rl|rl|}
\hline
                            & \multicolumn{2}{c|}{\textbf{DBLP QuAD}} & \multicolumn{2}{c|}{\textbf{CoyPuKGQA}} \\ 
\hline
OntoSCPrompt     & \multicolumn{1}{c}{78.2}       &       & \multicolumn{1}{c}{80.2}       &       \\ 
\hline
$\text{OntoSCPrompt}_{pre}$ & \multicolumn{1}{c}{84.6}       &   \textcolor{blue}{(+6.4)}    & \multicolumn{1}{c}{83.3}       &    \textcolor{blue}{(+3.1)}   \\ 
\hline
\end{tabular}
\label{tab:evaluation_4}
\end{table}

KGQA generalization across different KGs refers to the ability of a KGQA system to provide correct answers to questions across various KGs without extensive retraining. It involves transferring knowledge and reasoning capabilities from one KG to another. To assess the ability of OntoSCPrompt, we use the model pre-trained with a source KGQA dataset based on one KG and adapt to a target KGQA dataset based on another KG.

We find that the pre-trained variant improves the accuracy by \textbf{+6.4} on DBLP QuAD and \textbf{+3.1} on CoyPuKGQA respectively, showing that pre-training could bring significant performance gains. In particular on a domain-specific or even low-resource dataset, and facilitate generalization across multiple KGs, see Table~\ref{tab:evaluation_4}.


\subsection{Evaluation on Ontology-Guided Hybrid Prompts}

As discussed earlier, we construct KG ontology-guided hybrid prompts and introduce four continuous vectors to learn different aspects of the input. Thus, the trainable parameters correspond to the parameters of the base LLM $\Theta_m$ and the learnable vectors $\Theta_l$. However, the extent of the potential contribution of such a hybrid prompt to KGQA generalization remains uncertain. Thus, we conduct an ablation study to investigate its effect on the overall performance. First, we only fine-tune $\Theta_l$ with $\Theta_m$ fixed, i.e., prompt tuning (PT). Second, we fine-tune both $\Theta_m$ and $\Theta_l$ (i.e. PT+FT). The results are reported in Table~\ref{tab:evaluation_6}. It is evident that such a hybrid prompt contributes to the overall performance as only optimizing ontology-guided hybrid prompt results in 70.3\% on LC-QuAD 1.0 and 62.1\% on WebQSP. This highlights its importance in understanding the semantics of the underlying KG and adapting to unseen KG without extensive re-training.

\begin{table}[htb!]
\caption{The results of different fine-tuning strategies on LC-QuAD 1.0 and WebQSP.}
\footnotesize
\centering
\begin{tabular}{|c|l|l|}
\hline
                 & \textbf{WebQSP}  & \textbf{LC-QuAD 1.0} \\
\hline
PT   & 62.1  &   70.3    \\
PT + FT  & 73.8 \textcolor{blue}{(+11.7)}    &   79.1 \textcolor{blue}{(+8.8)}  \\
\hline
\end{tabular}
\label{tab:evaluation_6}
\end{table}

\subsection{Impact of Constrained Decoding}


\begin{figure}[h]
\centering
    \includegraphics[width=0.7\linewidth]{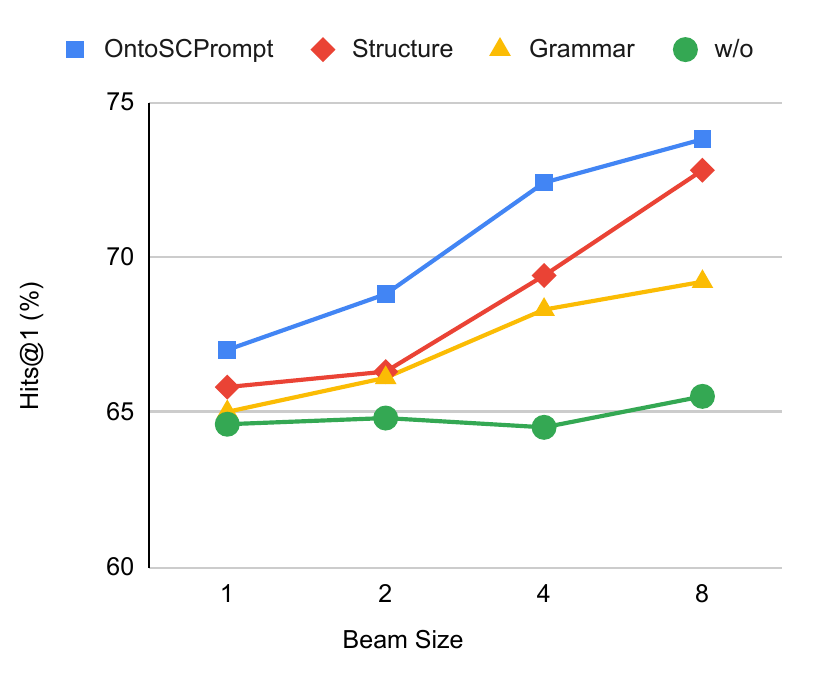}
    \caption{The performance of different decoding strategies on WebQSP under various beam sizes.}
    \label{fig:decoding}
\end{figure}

We perform an ablation analysis to examine the impact of each task-specific decoding technique on the performance. Note that regarding to the choice of WebQSP over LC-QuAD and CWQ for this evaluation, we have the following reasons: (1) in LC-QuAD 1.0, all test set structures are seen in the training set, leading to high accuracy in structure inference. (2) LC-QuAD 1.0 lacks constraint-based structures and is mostly multiple hops, while WebQSP and CWQ have more complex SPARQL structures. (3) WebQSP has 34\% unseen structures in the test set, compared to 12.4\% in CWQ, making it more challenging despite structural differences.
(4) Both CWQ and WebQSP are based on Freebase, making it unnecessary to evaluate both. The results are reported in Figure~\ref{fig:decoding}, indicating a rise in performance when employing each constrained decoding strategy as the beam size increases. OntoSCPrompt, equipped with all constrained decoding strategies, achieves superior performance, showing the effectiveness of our proposed decoding strategies for KGQA generalization.

\section{Related Work}

Our approach is in line with methodologies employing a two-stage architecture for semantic parsing tasks, akin to works such as Coarse2Fine~\cite{dong-lapata-2018-coarse}, STaG-QA~\cite{ravishankar-etal-2022-two}, and HGNet~\cite{10.1109/TKDE.2022.3207477}, where questions are first mapped to an initial outline and then filled in details later. However, they overlook condition expressions or constraints in SPARQL queries, whereas OntoSCPrompt's SPARQL structure representation is more comprehensive, enhancing KGQA generalization.
Most existing KGQA systems lack generalization as they are either typically tailored to a specific KG or focus only on within-a-KG generalization. While some methods have demonstrated limited ability to generalize across KGs, particularly in handling assertion heterogeneity between datasets such as WebQSP (Freebase) and MetaQA (Wikimovies), they fail to generalize across different schemas or topologies. LLMs also suffer from issues like hallucinations and factual inaccuracy when answering questions~\cite{10.1145/3571730}, specifically in the knowledge-intensive task KGQA~\cite{2023arXiv230307992T,2023arXiv230310368H}. Some studies~\cite{2023arXiv230911206W,baek-etal-2023-knowledge} resort to KG-augmented prompt, i.e., injecting question-related factual information (e.g., KG triples) into predefined templates. Hallucinations still remain in the context of KGQA generalization as they adapt to heterogeneous KGs. In this work, we integrate the ontology verbalized in a unified way into the prompt and guide LLMs to fulfil our task, facilitating reasoning over KG ontology.

\section{Conclusion}

We propose OntoSCPrompt, a novel KGQA model that can generalize across multiple KGs. Regarding similarities and differences between heterogeneous KGs, we employ a two-stage approach that separates semantic parsing from KG-dependent interactions. In the structure stage, the model predicts a SPARQL query sketch. In the content stage, the model fills the structure with KG-specific information. We also employ an ontology-guided hybrid prompt learning strategy where the KG ontology is integrated into the learning process of hybrid prompts, which we prove is effective in mitigating KG heterogeneity and facilitating KGQA generalization. Meanwhile, we propose several decoding strategies tailored to different stages to further improve the performance. We evaluate OntoSCPrompt on diverse KGQA datasets from different KGs for KGQA generalization. Experimental results demonstrated its effectiveness. 

\section*{Limitations}

Despite achieving state-of-the-art or competitive accuracy on KGQA benchmarks for generalization across multiple KGs, OntoSCPrompt still exhibits several limitations: \begin{enumerate*}[label=\arabic*)]

\item \textbf{Directionality of Relations}: The directionality of relations poses a challenge. For instance, in Freebase, the relation "capital" connects a country entity to a city entity as "\textlangle Germany\textrangle,\textlangle capital\textrangle,\textlangle Berlin\textrangle" while in other KGs it might be represented as "\textlangle Berlin\textrangle,\textlangle capital\textrangle,\textlangle Germany\textrangle". These variations in relation to directionality can result in errors.

\item \textbf{Differences in SPARQL Query Writing Style}: Different human annotators may annotate the same question using varying writing styles or SPARQL grammar instantiations. This diversity in annotation can restrict OntoSCPrompt's ability to generalize across multiple KGs. 

\item \textbf{Verbose Naming Convention}: Compared to the more concise conventions in DBpedia and Wikidata, in Freebase, the naming convention for schema elements tends to be verbose. This verbose approach leads to an explosive increase in the LLM's context length, posing challenges during training.

\end{enumerate*}




\bibliographystyle{IEEEtran}
\bibliography{main}

\begin{thebibliography}{10}
\providecommand{\url}[1]{#1}
\csname url@samestyle\endcsname
\providecommand{\newblock}{\relax}
\providecommand{\bibinfo}[2]{#2}
\providecommand{\BIBentrySTDinterwordspacing}{\spaceskip=0pt\relax}
\providecommand{\BIBentryALTinterwordstretchfactor}{4}
\providecommand{\BIBentryALTinterwordspacing}{\spaceskip=\fontdimen2\font plus
\BIBentryALTinterwordstretchfactor\fontdimen3\font minus \fontdimen4\font\relax}
\providecommand{\BIBforeignlanguage}[2]{{%
\expandafter\ifx\csname l@#1\endcsname\relax
\typeout{** WARNING: IEEEtran.bst: No hyphenation pattern has been}%
\typeout{** loaded for the language `#1'. Using the pattern for}%
\typeout{** the default language instead.}%
\else
\language=\csname l@#1\endcsname
\fi
#2}}
\providecommand{\BIBdecl}{\relax}
\BIBdecl

\bibitem{kapanipathi-etal-2021-leveraging}
P.~Kapanipathi, I.~Abdelaziz, S.~Ravishankar, S.~Roukos, A.~Gray, R.~Fernandez~Astudillo, M.~Chang, C.~Cornelio, S.~Dana, A.~Fokoue, D.~Garg, A.~Gliozzo, S.~Gurajada, H.~Karanam, N.~Khan, D.~Khandelwal, Y.-S. Lee, Y.~Li, F.~Luus, N.~Makondo, N.~Mihindukulasooriya, T.~Naseem, S.~Neelam, L.~Popa, R.~Gangi~Reddy, R.~Riegel, G.~Rossiello, U.~Sharma, G.~P.~S. Bhargav, and M.~Yu, ``Leveraging {A}bstract {M}eaning {R}epresentation for knowledge base question answering,'' in \emph{Findings of the Association for Computational Linguistics: ACL-IJCNLP 2021}, C.~Zong, F.~Xia, W.~Li, and R.~Navigli, Eds.\hskip 1em plus 0.5em minus 0.4em\relax Online: Association for Computational Linguistics, Aug. 2021, pp. 3884--3894.

\bibitem{10.1145/2588555.2610525}
L.~Zou, R.~Huang, H.~Wang, J.~X. Yu, W.~He, and D.~Zhao, ``Natural language question answering over rdf: a graph data driven approach,'' in \emph{Proceedings of the 2014 ACM SIGMOD International Conference on Management of Data}, ser. SIGMOD '14.\hskip 1em plus 0.5em minus 0.4em\relax New York, NY, USA: Association for Computing Machinery, 2014, p. 313–324.

\bibitem{10.1145/3357384.3358026}
S.~Vakulenko, J.~D. Fernandez~Garcia, A.~Polleres, M.~de~Rijke, and M.~Cochez, ``Message passing for complex question answering over knowledge graphs,'' in \emph{Proceedings of the 28th ACM International Conference on Information and Knowledge Management}, ser. CIKM '19.\hskip 1em plus 0.5em minus 0.4em\relax New York, NY, USA: Association for Computing Machinery, 2019, p. 1431–1440.

\bibitem{10.1145/3442381.3449992}
Y.~Gu, S.~Kase, M.~Vanni, B.~Sadler, P.~Liang, X.~Yan, and Y.~Su, ``Beyond i.i.d.: Three levels of generalization for question answering on knowledge bases,'' in \emph{Proceedings of the Web Conference 2021}, ser. WWW '21.\hskip 1em plus 0.5em minus 0.4em\relax New York, NY, USA: Association for Computing Machinery, 2021, p. 3477–3488.

\bibitem{gu-etal-2023-dont}
Y.~Gu, X.~Deng, and Y.~Su, ``Don{'}t generate, discriminate: A proposal for grounding language models to real-world environments,'' in \emph{Proceedings of the 61st Annual Meeting of the Association for Computational Linguistics (Volume 1: Long Papers)}, A.~Rogers, J.~Boyd-Graber, and N.~Okazaki, Eds.\hskip 1em plus 0.5em minus 0.4em\relax Toronto, Canada: Association for Computational Linguistics, Jul. 2023, pp. 4928--4949.

\bibitem{shu-yu-2024-distribution}
Y.~Shu and Z.~Yu, ``Distribution shifts are bottlenecks: Extensive evaluation for grounding language models to knowledge bases,'' in \emph{Proceedings of the 18th Conference of the European Chapter of the Association for Computational Linguistics: Student Research Workshop}, N.~Falk, S.~Papi, and M.~Zhang, Eds.\hskip 1em plus 0.5em minus 0.4em\relax St. Julian{'}s, Malta: Association for Computational Linguistics, Mar. 2024, pp. 71--88.

\bibitem{sun-etal-2018-open}
H.~Sun, B.~Dhingra, M.~Zaheer, K.~Mazaitis, R.~Salakhutdinov, and W.~Cohen, ``Open domain question answering using early fusion of knowledge bases and text,'' in \emph{Proceedings of the 2018 Conference on Empirical Methods in Natural Language Processing}, E.~Riloff, D.~Chiang, J.~Hockenmaier, and J.~Tsujii, Eds.\hskip 1em plus 0.5em minus 0.4em\relax Brussels, Belgium: Association for Computational Linguistics, Oct.-Nov. 2018, pp. 4231--4242.

\bibitem{saxena-etal-2020-improving}
A.~Saxena, A.~Tripathi, and P.~Talukdar, ``Improving multi-hop question answering over knowledge graphs using knowledge base embeddings,'' in \emph{Proceedings of the 58th Annual Meeting of the Association for Computational Linguistics}, D.~Jurafsky, J.~Chai, N.~Schluter, and J.~Tetreault, Eds.\hskip 1em plus 0.5em minus 0.4em\relax Online: Association for Computational Linguistics, Jul. 2020, pp. 4498--4507.

\bibitem{qiao-etal-2022-exploiting}
Z.~Qiao, W.~Ye, T.~Zhang, T.~Mo, W.~Li, and S.~Zhang, ``Exploiting hybrid semantics of relation paths for multi-hop question answering over knowledge graphs,'' in \emph{Proceedings of the 29th International Conference on Computational Linguistics}.\hskip 1em plus 0.5em minus 0.4em\relax Gyeongju, Republic of Korea: International Committee on Computational Linguistics, Oct. 2022, pp. 1813--1822.

\bibitem{mavromatis-karypis-2022-rearev}
C.~Mavromatis and G.~Karypis, ``{R}ea{R}ev: Adaptive reasoning for question answering over knowledge graphs,'' in \emph{Findings of the Association for Computational Linguistics: EMNLP 2022}, Y.~Goldberg, Z.~Kozareva, and Y.~Zhang, Eds.\hskip 1em plus 0.5em minus 0.4em\relax Abu Dhabi, United Arab Emirates: Association for Computational Linguistics, Dec. 2022, pp. 2447--2458.

\bibitem{10.1145/1376616.1376746}
K.~Bollacker, C.~Evans, P.~Paritosh, T.~Sturge, and J.~Taylor, ``Freebase: a collaboratively created graph database for structuring human knowledge,'' in \emph{Proceedings of the 2008 ACM SIGMOD International Conference on Management of Data}, ser. SIGMOD '08.\hskip 1em plus 0.5em minus 0.4em\relax New York, NY, USA: Association for Computing Machinery, 2008, p. 1247–1250.

\bibitem{10.5555/3504035.3504780}
Y.~Zhang, H.~Dai, Z.~Kozareva, A.~J. Smola, and L.~Song, ``Variational reasoning for question answering with knowledge graph,'' in \emph{Proceedings of the Thirty-Second AAAI Conference on Artificial Intelligence and Thirtieth Innovative Applications of Artificial Intelligence Conference and Eighth AAAI Symposium on Educational Advances in Artificial Intelligence}, ser. AAAI'18/IAAI'18/EAAI'18.\hskip 1em plus 0.5em minus 0.4em\relax AAAI Press, 2018.

\bibitem{2023arXiv230310368H}
N.~{Hu}, Y.~{Wu}, G.~{Qi}, D.~{Min}, J.~{Chen}, J.~Z. {Pan}, and Z.~{Ali}, ``{An Empirical Study of Pre-trained Language Models in Simple Knowledge Graph Question Answering},'' \emph{arXiv e-prints}, p. arXiv:2303.10368, Mar. 2023.

\bibitem{2023arXiv230307992T}
Y.~{Tan}, D.~{Min}, Y.~{Li}, W.~{Li}, N.~{Hu}, Y.~{Chen}, and G.~{Qi}, ``{Can ChatGPT Replace Traditional KBQA Models? An In-depth Analysis of the Question Answering Performance of the GPT LLM Family},'' \emph{arXiv e-prints}, p. arXiv:2303.07992, Mar. 2023.

\bibitem{2023arXiv230911206W}
Y.~{Wu}, N.~{Hu}, S.~{Bi}, G.~{Qi}, J.~{Ren}, A.~{Xie}, and W.~{Song}, ``{Retrieve-Rewrite-Answer: A KG-to-Text Enhanced LLMs Framework for Knowledge Graph Question Answering},'' \emph{arXiv e-prints}, p. arXiv:2309.11206, Sep. 2023.

\bibitem{10.1145/3571730}
Z.~Ji, N.~Lee, R.~Frieske, T.~Yu, D.~Su, Y.~Xu, E.~Ishii, Y.~J. Bang, A.~Madotto, and P.~Fung, ``Survey of hallucination in natural language generation,'' \emph{ACM Comput. Surv.}, vol.~55, no.~12, mar 2023.

\bibitem{DBLP:conf/esws/KlagerP23}
\BIBentryALTinterwordspacing
G.~Klager and A.~Polleres, ``Is {GPT} fit for kgqa? - preliminary results,'' in \emph{Joint Proceedings of the Second International Workshop on Knowledge Graph Generation From Text and the First International BiKE Challenge co-located with 20th Extended Semantic Conference {(ESWC} 2023), Hersonissos, Greece, May 29th, 2023}, ser. {CEUR} Workshop Proceedings, S.~Tiwari, N.~Mihindukulasooriya, F.~Osborne, D.~Kontokostas, J.~D'Souza, M.~Kejriwal, and E.~Marx, Eds., vol. 3447.\hskip 1em plus 0.5em minus 0.4em\relax CEUR-WS.org, 2023, pp. 171--191. [Online]. Available: \url{https://ceur-ws.org/Vol-3447/Text2KG\_Paper\_11.pdf}
\BIBentrySTDinterwordspacing

\bibitem{ravishankar-etal-2022-two}
S.~Ravishankar, D.~Thai, I.~Abdelaziz, N.~Mihindukulasooriya, T.~Naseem, P.~Kapanipathi, G.~Rossiello, and A.~Fokoue, ``A two-stage approach towards generalization in knowledge base question answering,'' in \emph{Findings of the Association for Computational Linguistics: EMNLP 2022}, Y.~Goldberg, Z.~Kozareva, and Y.~Zhang, Eds.\hskip 1em plus 0.5em minus 0.4em\relax Abu Dhabi, United Arab Emirates: Association for Computational Linguistics, Dec. 2022, pp. 5571--5580.

\bibitem{10.1145/3589292}
Z.~Gu, J.~Fan, N.~Tang, L.~Cao, B.~Jia, S.~Madden, and X.~Du, ``Few-shot text-to-sql translation using structure and content prompt learning,'' \emph{Proc. ACM Manag. Data}, vol.~1, no.~2, jun 2023.

\bibitem{10.1007/978-3-540-76298-0_52}
S.~Auer, C.~Bizer, G.~Kobilarov, J.~Lehmann, R.~Cyganiak, and Z.~Ives, ``Dbpedia: A nucleus for a web of open data,'' in \emph{The Semantic Web}, K.~Aberer, K.-S. Choi, N.~Noy, D.~Allemang, K.-I. Lee, L.~Nixon, J.~Golbeck, P.~Mika, D.~Maynard, R.~Mizoguchi, G.~Schreiber, and P.~Cudr{\'e}-Mauroux, Eds.\hskip 1em plus 0.5em minus 0.4em\relax Berlin, Heidelberg: Springer Berlin Heidelberg, 2007, pp. 722--735.

\bibitem{DBLP:conf/birws/BanerjeeAUB23}
D.~Banerjee, S.~Awale, R.~Usbeck, and C.~Biemann, ``Dblp-quad: {A} question answering dataset over the {DBLP} scholarly knowledge graph,'' in \emph{Proceedings of the 13th International Workshop on Bibliometric-enhanced Information Retrieval co-located with 45th European Conference on Information Retrieval {(ECIR} 2023), Dublin, Ireland, April 2nd, 2023}, ser. {CEUR} Workshop Proceedings, I.~Frommholz, P.~Mayr, G.~Cabanac, S.~Verberne, and J.~Brennan, Eds., vol. 3617.\hskip 1em plus 0.5em minus 0.4em\relax CEUR-WS.org, 2023, pp. 37--51.

\bibitem{DBLP:conf/semweb/00010U23}
L.~Jiang, X.~Yan, and R.~Usbeck, ``A structure and content prompt-based method for knowledge graph question answering over scholarly data,'' in \emph{Joint Proceedings of Scholarly {QALD} 2023 and SemREC 2023 co-located with 22nd International Semantic Web Conference {ISWC} 2023, Athens, Greece, November 6-10, 2023}, ser. {CEUR} Workshop Proceedings, D.~Banerjee, R.~Usbeck, N.~Mihindukulasooriya, G.~Singh, R.~Mutharaju, and P.~Kapanipathi, Eds., vol. 3592.\hskip 1em plus 0.5em minus 0.4em\relax CEUR-WS.org, 2023.

\bibitem{text2kgbench}
N.~Mihindukulasooriya, S.~Tiwari, C.~F. Enguix, and K.~Lata, ``Text2kgbench: A benchmark for ontology-driven knowledge graph generation from text,'' in \emph{The Semantic Web – ISWC 2023: 22nd International Semantic Web Conference, Athens, Greece, November 6–10, 2023, Proceedings, Part II}.\hskip 1em plus 0.5em minus 0.4em\relax Berlin, Heidelberg: Springer-Verlag, 2023, p. 247–265.

\bibitem{zheng-lapata-2021-compositional-generalization}
H.~Zheng and M.~Lapata, ``Compositional generalization via semantic tagging,'' in \emph{Findings of the Association for Computational Linguistics: EMNLP 2021}, M.-F. Moens, X.~Huang, L.~Specia, and S.~W.-t. Yih, Eds.\hskip 1em plus 0.5em minus 0.4em\relax Punta Cana, Dominican Republic: Association for Computational Linguistics, Nov. 2021, pp. 1022--1032.

\bibitem{leblond-etal-2021-machine}
R.~Leblond, J.-B. Alayrac, L.~Sifre, M.~Pislar, L.~Jean-Baptiste, I.~Antonoglou, K.~Simonyan, and O.~Vinyals, ``Machine translation decoding beyond beam search,'' in \emph{Proceedings of the 2021 Conference on Empirical Methods in Natural Language Processing}, M.-F. Moens, X.~Huang, L.~Specia, and S.~W.-t. Yih, Eds.\hskip 1em plus 0.5em minus 0.4em\relax Online and Punta Cana, Dominican Republic: Association for Computational Linguistics, Nov. 2021, pp. 8410--8434.

\bibitem{fan-etal-2018-controllable}
A.~Fan, D.~Grangier, and M.~Auli, ``Controllable abstractive summarization,'' in \emph{Proceedings of the 2nd Workshop on Neural Machine Translation and Generation}, A.~Birch, A.~Finch, T.~Luong, G.~Neubig, and Y.~Oda, Eds.\hskip 1em plus 0.5em minus 0.4em\relax Melbourne, Australia: Association for Computational Linguistics, Jul. 2018, pp. 45--54.

\bibitem{baranowski-hochgeschwender-2021-grammar2}
A.~Baranowski and N.~Hochgeschwender, ``Grammar-constrained neural semantic parsing with {LR} parsers,'' in \emph{Findings of the Association for Computational Linguistics: ACL-IJCNLP 2021}, C.~Zong, F.~Xia, W.~Li, and R.~Navigli, Eds.\hskip 1em plus 0.5em minus 0.4em\relax Online: Association for Computational Linguistics, Aug. 2021, pp. 1275--1279.

\bibitem{geng-etal-2023-grammar}
S.~Geng, M.~Josifoski, M.~Peyrard, and R.~West, ``Grammar-constrained decoding for structured {NLP} tasks without finetuning,'' in \emph{Proceedings of the 2023 Conference on Empirical Methods in Natural Language Processing}, H.~Bouamor, J.~Pino, and K.~Bali, Eds.\hskip 1em plus 0.5em minus 0.4em\relax Singapore: Association for Computational Linguistics, Dec. 2023.

\bibitem{DBLP:conf/acl/ZhangZY000C22}
J.~Zhang, X.~Zhang, J.~Yu, J.~Tang, J.~Tang, C.~Li, and H.~Chen, ``Subgraph retrieval enhanced model for multi-hop knowledge base question answering,'' in \emph{Proceedings of the 60th Annual Meeting of the Association for Computational Linguistics (Volume 1: Long Papers), {ACL} 2022, Dublin, Ireland, May 22-27, 2022}, S.~Muresan, P.~Nakov, and A.~Villavicencio, Eds.\hskip 1em plus 0.5em minus 0.4em\relax Association for Computational Linguistics, 2022, pp. 5773--5784.

\bibitem{jiang-etal-2023-reasoninglm}
J.~Jiang, K.~Zhou, X.~Zhao, Y.~Li, and J.-R. Wen, ``{R}easoning{LM}: Enabling structural subgraph reasoning in pre-trained language models for question answering over knowledge graph,'' in \emph{Proceedings of the 2023 Conference on Empirical Methods in Natural Language Processing}, H.~Bouamor, J.~Pino, and K.~Bali, Eds.\hskip 1em plus 0.5em minus 0.4em\relax Singapore: Association for Computational Linguistics, Dec. 2023, pp. 3721--3735.

\bibitem{sun-etal-2019-pullnet}
H.~Sun, T.~Bedrax-Weiss, and W.~Cohen, ``{P}ull{N}et: Open domain question answering with iterative retrieval on knowledge bases and text,'' in \emph{Proceedings of the 2019 Conference on Empirical Methods in Natural Language Processing and the 9th International Joint Conference on Natural Language Processing (EMNLP-IJCNLP)}, K.~Inui, J.~Jiang, V.~Ng, and X.~Wan, Eds.\hskip 1em plus 0.5em minus 0.4em\relax Hong Kong, China: Association for Computational Linguistics, Nov. 2019, pp. 2380--2390.

\bibitem{10.1109/TKDE.2022.3207477}
Y.~Chen, H.~Li, G.~Qi, T.~Wu, and T.~Wang, ``Outlining and filling: Hierarchical query graph generation for answering complex questions over knowledge graphs,'' \emph{IEEE Trans. on Knowl. and Data Eng.}, vol.~35, no.~8, p. 8343–8357, aug 2023.

\bibitem{guo-etal-2022-longt5}
M.~Guo, J.~Ainslie, D.~Uthus, S.~Ontanon, J.~Ni, Y.-H. Sung, and Y.~Yang, ``{L}ong{T}5: {E}fficient text-to-text transformer for long sequences,'' in \emph{Findings of the Association for Computational Linguistics: NAACL 2022}, M.~Carpuat, M.-C. de~Marneffe, and I.~V. Meza~Ruiz, Eds.\hskip 1em plus 0.5em minus 0.4em\relax Seattle, United States: Association for Computational Linguistics, Jul. 2022, pp. 724--736.

\bibitem{zhang-etal-2022-subgraph}
J.~Zhang, X.~Zhang, J.~Yu, J.~Tang, J.~Tang, C.~Li, and H.~Chen, ``Subgraph retrieval enhanced model for multi-hop knowledge base question answering,'' in \emph{Proceedings of the 60th Annual Meeting of the Association for Computational Linguistics (Volume 1: Long Papers)}, S.~Muresan, P.~Nakov, and A.~Villavicencio, Eds.\hskip 1em plus 0.5em minus 0.4em\relax Dublin, Ireland: Association for Computational Linguistics, May 2022, pp. 5773--5784.

\bibitem{hartmann-marx-soru-2018}
A.-K. Hartmann, E.~Marx, and T.~Soru, ``Generating a large dataset for neural question answering over the {DB}pedia knowledge base,'' 2018.

\bibitem{dong-lapata-2018-coarse}
L.~Dong and M.~Lapata, ``Coarse-to-fine decoding for neural semantic parsing,'' in \emph{Proceedings of the 56th Annual Meeting of the Association for Computational Linguistics (Volume 1: Long Papers)}, I.~Gurevych and Y.~Miyao, Eds.\hskip 1em plus 0.5em minus 0.4em\relax Melbourne, Australia: Association for Computational Linguistics, Jul. 2018, pp. 731--742.

\bibitem{baek-etal-2023-knowledge}
J.~Baek, A.~F. Aji, and A.~Saffari, ``Knowledge-augmented language model prompting for zero-shot knowledge graph question answering,'' in \emph{Proceedings of the 1st Workshop on Natural Language Reasoning and Structured Explanations (NLRSE)}, B.~Dalvi~Mishra, G.~Durrett, P.~Jansen, D.~Neves~Ribeiro, and J.~Wei, Eds.\hskip 1em plus 0.5em minus 0.4em\relax Toronto, Canada: Association for Computational Linguistics, Jun. 2023, pp. 78--106.

\end{thebibliography}





\end{document}